\documentclass[conference]{IEEEtran}
\usepackage{cite}

\usepackage{latexsym}
\usepackage{amssymb}
\usepackage{amsmath}
\usepackage{amsthm}
\usepackage{booktabs}
\usepackage{enumitem}
\usepackage{graphicx}
\usepackage{soul}
\usepackage{hyperref}
\usepackage{pgfplots}

\usepackage{float} 
\usepackage{array} 

\usepackage{colortbl} 
\usepackage{soul} 
\usepackage{xcolor} 
\soulregister\cite7 
\usepackage{caption}
\usepackage{subcaption}
\definecolor{blue}{RGB}{31, 119, 180}
\definecolor{orange}{RGB}{255, 127, 14}
\definecolor{green}{RGB}{44, 160, 44}
\definecolor{red}{RGB}{214, 39, 40}
\definecolor{purple}{RGB}{148, 103, 189}
\definecolor{brown}{RGB}{140, 86, 75}
\definecolor{pink}{RGB}{227, 119, 194}
\definecolor{gray}{RGB}{127, 127, 127}
\definecolor{yellow}{RGB}{188, 189, 34}
\definecolor{teal}{RGB}{23, 190, 207}

\usepackage{multirow}
\usepackage{tikz}
\usepackage{tabularx}
\usepackage{longtable}
\usepackage{pdflscape}
\usepackage{xltabular} 
\usepackage{ragged2e}  
\newcolumntype{L}{>{\RaggedRight\hspace{0pt}}X}
\newcolumntype{C}{>{\centering\arraybackslash}X}
\usepackage{enumitem}
\usepackage{booktabs}

\newcommand*\colourcheck[1]{%
  \expandafter\newcommand\csname #1check\endcsname{\textcolor{#1}{\ding{52}}}%
}

\newcommand{\BibTeX}{B\kern-.05em{\sc i\kern-.025em b}\kern-.08em\TeX}

\pgfplotsset{compat=1.11,
        /pgfplots/ybar legend/.style={
        /pgfplots/legend image code/.code={%
        \draw[##1,/tikz/.cd,bar width=3pt,yshift=-0.2em,bar shift=0pt]
                plot coordinates {(0cm,0.8em)};},
},
}

\usepackage{xspace}
\newcommand{\ourname}{\texttt{MariData}\xspace} 

\begin{document}


\title{\ourname: One-Step Unpaired Image Translation for Maritime Environments}

\author{Santeri~Henriksson\textsuperscript{*}, Mehdi~Asadi\textsuperscript{*}, Amin Majd{*}, Juha Kalliovaara{*}\\[1ex]
\textsuperscript{*}AIS Lab, Turku University of Applied Sciences, Turku, Finland
}

\maketitle

\begin{abstract}
The development on robust perception systems for Maritime Autonomous Surface Ships (MASS) is heavily constrained by the scarcity of diverse training data, particularly for adverse weather and low-light conditions. Because collecting paired images in dynamic maritime environments is physically impossible, synthetic data generation via unpaired image-to-image translation offers a critical solution. However, existing generative models suffer from failing to preserve the fine structural details of small navigational objects due to latent compression bottlenecks. In this paper, we introduce a framework for generating synthetic maritime data using CycleGAN-turbo, a one-step unpaired translation architecture. By incorporating zero-convolution skip connections to bypass the Variational Autoencoder (VAE) bottleneck, our approach explicitly preserves small object details (e.g., distant vessels and sea marks) during translation. We compiled a dataset of 7,000 maritime images to train and evaluate models for Day-to-Foggy, Day-to-Sunset, and Day-to-Night domain translations. Qualitative evaluations and variable-strength inference studies demonstrate that our method effectively synthesizes realistic atmospheric conditions while maintaining the underlying semantic structure of the scene. The Day-to-Foggy and Day-to-Sunset models exhibit great structural retention, whereas the Day-to-Night model highlights the challenge of semantic hallucination, such as generating artificial coastal lights, induced by unbalanced training distributions. Ultimately, this work establishes an efficient, structure-aware data synthesis pipeline that directly addresses the data scarcity bottleneck in autonomous maritime navigation.

\end{abstract}

\section{Introduction}
\label{sec:introduction}

The deployment of Maritime Autonomous Surface Ships (MASS) relies fundamentally on robust and reliable perception systems capable of navigating complex, real-world environments. While significant progress has been made under benign weather and lighting conditions, the performance of these systems is heavily constrained. Specifically, there is a severe scarcity of diverse training data representing adverse weather like fog and low-light scenarios such as sunset or night. Unlike autonomous driving on roads, the stochastic nature of maritime environments makes collecting perfectly aligned, paired images of the exact same scene in different domains virtually impossible. Dynamic wave patterns and floating objects ensure the visual scene is constantly changing minute by minute. Consequently, data scarcity remains a critical bottleneck for developing reliable all-weather maritime navigation systems.

To overcome the lack of paired real-world data, synthetic data generation via Generative Adversarial Networks (GANs) and recent Diffusion models has emerged as a promising solution. Unpaired image-to-image translation frameworks, such as CycleGAN \cite{zhu2017cyclegan}, have demonstrated the ability to transfer styles between unordered domains. However, standard generative models face severe limitations when applied directly to maritime datasets. Traditional GANs often suffer from structural distortions and fail to consistently generate photorealistic surface textures. Conversely, while state-of-the-art latent diffusion models produce high-fidelity images, they suffer from high inference latency and rely on a restrictive Variational Autoencoder (VAE) bottleneck. This spatial compression process inherently discards high-frequency details, causing small but navigationally critical objects like distant vessels or buoys to be erased or heavily distorted.

Addressing the dual challenges of latency and structural degradation, we propose a novel framework for synthetic maritime data generation utilizing CycleGAN-turbo \cite{parmar2024cycleganturbo}, a one-step unpaired translation architecture. By incorporating zero-convolution skip connections that bypass the restrictive VAE bottleneck, our approach directly preserves the fine-grained structural details of small navigational objects. This allows the model to faithfully retain critical semantic features while effectively adapting the global atmospheric context. In this paper, we focus strictly on generating realistic adverse weather and lighting conditions from standard daytime images. Through this methodology, we can fundamentally prioritize maintaining the underlying semantic integrity of the scene throughout the translation process. 

Specifically, the contributions of this work are threefold, addressing the core issues of data scarcity and synthetic generation limitations. First, we adapt the CycleGAN-turbo architecture for maritime environments to achieve one-step, structure-aware image translation, explicitly solving the small-object omission problem caused by VAE bottlenecks. Second, we compile and utilize a dataset of 7,000 maritime images to actively train and evaluate distinct domain translation models for Day-to-Foggy, Day-to-Sunset, and Day-to-Night scenarios. Third, we provide comprehensive qualitative evaluations and variable-strength inference studies across all models. These studies demonstrate the effectiveness of our approach in maintaining the structural integrity of navigational aids throughout translation. They also provide key analysis regarding the model's limitations concerning semantic hallucinations in imbalanced domains.

Ultimately, this work establishes an efficient and robust data synthesis pipeline designed specifically for challenging domain transfers. By directly addressing the critical data scarcity bottleneck, we supply a method for computationally generating otherwise unattainable training images. This significantly reduces the reliance on physically collecting rare, dangerous, or perfectly paired multi-domain real-world datasets. Furthermore, explicitly retaining the structural integrity of minute objects guarantees the synthetic data remains viably useful for perception algorithm training. Through this framework, we pave the direct technical way for the deployment of far more resilient autonomous maritime navigation systems.

\label{sec:introduction}

\section{Related work}
\label{Related work}
\subsection{Maritime Perception and Data Scarcity}
The development of robust perception systems for Maritime Autonomous Surface Ships (MASS) is constrained by the availability of diverse training data. Early datasets primarily focused on specific tasks such as search and rescue or traffic monitoring. For instance, the Singapore Maritime Dataset (SMD) provides on-shore video for traffic analysis, while SeaDronesSee and MOBDrone primarily target detecting humans and life rafts in open water for rescue operations. \cite{prasad2015smd} \cite{varga2022seadronessee} \cite{ciampi2022mobdrone}

However, these datasets often lack the environmental complexity required for archipelago navigation. Addressing this, Asadi et al. introduced the DIANA dataset, which distinguishes between navigationally critical classes (e.g., sea marks, motorboats) within complex littoral environments. \cite{asadi2025diana} Despite these advances, a critical limitation across real-world maritime datasets is the scarcity of adverse atmospheric conditions (e.g., dense fog, night). Collecting paired data where the exact same scene is captured in both "Day" and "Night" conditions is physically impossible in maritime environments due to the stochastic nature of waves and floating objects. This reliance on rare, unpaired real-world data necessitates the use of synthetic data generation.

\subsection{Generative Image-to-Image Translation}
To synthesize training data without paired examples, researchers have turned to Generative Adversarial Networks (GANs). The field shifted from supervised methods like Pix2Pix, which strictly require aligned image pairs \cite{isola2017pix2pix}, to unsupervised frameworks. CycleGAN, introduced by Zhu et al., solved the paired data problem by enforcing a cycle-consistency loss (\(F(G(x))\approx x\)), allowing translation between unordered domains (e.g., Summer \(\leftrightarrow\) Winter). \cite{zhu2017cyclegan}

Standard GAN architectures often suffer from training instability and "mode collapse", failing to generate high-fidelity textures. Newer approaches like Contrastive Unpaired Translation (CUT) attempted to improve structure preservation by maximizing mutual information between input and output patches rather than full-image reconstruction \cite{park2020cut}, yet they often lack the photorealistic texture synthesis capabilities of modern diffusion models.

\subsection{The Diffusion Paradigm and the "Small Object" Gap}
Recent advancements have seen Diffusion Models (e.g., Stable Diffusion) surpass GANs in image quality. Conditional adaptations like ControlNet allow for spatial guidance using edge maps. \cite{zhang2023controlnet} However, these models introduce two significant bottlenecks for maritime application:
\begin{enumerate}
    \item \textbf{Latency:} The iterative denoising process (often 20-50 steps) is computationally expensive, limiting rapid data generation.
    \item \textbf{The VAE Bottleneck:} Latent diffusion models rely on Variational Autoencoders (VAEs) that spatially compress images, often by 8x (DALL-E) or even 48x (Stable Diffusion). In this compression, small objects like sea marks or distant vessels are often discarded as high-frequency noise and fail to reconstruct in the generated image.
\end{enumerate}

\subsection{The Solution: One-Step Distillation}
To bridge the gap between GAN speed and Diffusion quality, recent research has focused on Adversarial Diffusion Distillation (ADD). Parmar et al. introduced CycleGAN-turbo, a one-step translation architecture that adapts pre-trained diffusion models to new domains using adversarial objectives. \cite{parmar2024cycleganturbo} Crucially, unlike standard latent diffusion, CycleGAN-turbo incorporates skip connections and zero-convolutions that bypass the VAE bottleneck.

\textbf{The Pivot:} While CycleGAN-turbo has proven effective for general domain transfer, it's application to specific constraints of maritime autonomy, specifically the preservation of minute navigational aids during lighting changes (Day-To-Sunset) and weather conditions, remains underexplored. Our work utilizes this architecture to create a pipeline for generating adverse weather maritime data that retains the structural integrity of small objects, directly addressing the data scarcity identified in the USVA project scope.
\section{methods}
\label{methods}

To enable the translation of maritime scenes across different atmospheric domain, a diverse dataset of unpaired images was compiled. The primary sources for this imagery were Unsplash and Kaggle, selected to ensure a wide variety of vessel types, sea states, and lighting conditions. Data collection was automated using a web scraping pipeline targeting specific semantic categories such as "maritime sunrise", "maritime day", and "maritime night".

The raw dataset underwent manual curation process to define environmental domains, Initially, the research design considered five categories: \verb|day|, \verb|night|, \verb|foggy|, \verb|sunset/sunrise|, and \verb|stormy|. However, the \verb|stormy| category was excluded during the curation phase due to excess intra-class variance, where visual features ranged inconsistent extremes (e.g., calm grey waters vs. violent whitecapes), which proved detrimental to model convergence.

\subsection{Annotation Infrastructure}
Annotation was performed using a local deployment of Label Studio \cite{labelStudio} on an Ubuntu 22.04 environment. To handle the privacy and latency requirements of high-resolution image datasets, the infrastructure was configured for local file serving, bypassing cloud storage requirements.

The labeling workflow enforced a strict single-choice classification policy. Annotator was presented with raw images and required to assign a global atmospheric label (\verb|day|, \verb|night|, \verb|foggy|, \verb|sunset/sunrise|, and \verb|stormy|). The resulting annotations were exported in CSV format and processed via a custom Python script to sanitize file paths, converting absolute system paths into relative paths compatible with the training pipeline.

\subsection{Data Preprocessing and Partitioning}
Prior to training, all imagery was subjected to a standardized preprocessing pipeline to ensure tensor compatibility with the CycleGAN-turbo architecture.
\begin{enumerate}
    \item \textbf{Resizing and Cropping:} Input images were resized using bicubic interpolation to 286x286 pixels. During training, a random crop of 256x256 pixels was applied to introduce spatial variance and prevent overfitting. For inference and testing, a center crop of 256x256 was utilized to ensure deterministic evaluation.
    \item \textbf{Dimensional Constraints:}
The resolution was strictly constrained to multiples of 16 (e.g., \(256 = 16 \times 16\)) to align with the downsampling stride of the convolutional encoder-decoder architecture.
    \item \textbf{Unpaired Splitting:}
The dataset was partitioned into an unpaired structure required by the CycleGAN framework (\verb|train_A|, \verb|train_B|, \verb|test_A|, \verb|test_B|). A 90/10 split ratio was used, reserving 10 percent of the curated data for evaluation purposes.
\end{enumerate}

\subsection{Model Architecture}
This research utilizes the CycleGAN-turbo framework, a one-step image-to-image translation architecture that integrates the cycle-consistency constraints of Generative Adversarial Networks (GANs) with the high-fidelity synthesis capabilities of diffusion models. \cite{parmar2024cycleganturbo}

Unlike traditional GANs which often rely on on pixel-level losses that can lead to blurring, CycleGAN-turbo uses a pre-trained Stable Diffusion Turbo backbone. This choice was driven by the project's requirement to prioritize global atmospheric adaptation, accurately rendering complex weather phenomena like fog density and lighting shifts while also retaining small object detail. The model learns two simultaneous mappings: 
\begin{enumerate}
    \item \textbf{Forward} (\(G_{A \to B}\)): Source Domain (e.q., \verb|day|) \(\rightarrow\) Target Domain (e.g., \verb|night|).
    \item \textbf{Inverse} (\(G_{B \to A}\)): Target Domain \(\rightarrow\) Source Domain. 
\end{enumerate}
The preservation of semantic content (e.g., vessel structure, horizon line) is enforced via Cycle Consistency Loss, which penalizes differences between the original image and the image reconstructed after a round-trip translation (\(A \to B \to A\)).

\subsection{Training Implementation}
The model was trained on a single NVIDIA\textsuperscript{\textregistered} RTX 4090 GPU. Attempts to scale training to a multi-GPU cluster (DGX with Tesla V100s) were deprecated due to driver incompatabilities between the required CUDA version (11.7+) for PyTorch 2.0.1 and the available cluster environment (CUDA 11.4).

\subsubsection{Environment and Optimization}
The training environment was containerized using Conda with Python 3.10. To accommodate the memory requirements of the diffusion backbone on a single GPU, several optimization techniques were used via the Hugging Face Accelerate library:
\begin{itemize}
    \item \textbf{Mixed Precision (FP16):} Enabled to reduce VRAM usage and accelerate tensor operations.
    \item \textbf{xFormers Attention:} Utilized to implement memory-efficient attention mechanisms.
    \item \textbf{Gradient Checkpointing:} Applied to trade computational overhead for memory savings, preventing Out-Of-Memory (OOM) errors.
    \item \textbf{Monkey-Patching:} Custom compatability shims were put into the training scripts to address deprecated functions in the \verb|huggingface_hub| and \verb|transformers| libraries, ensuring the correct loading of the pre-trained SD-Turbo weights.
\end{itemize}

\subsubsection{Hyperparameters}
The training process utilized the following hyperparameter configuration:
\begin{itemize}
    \item \textbf{Learning Rate:} \(1e-5\)
    \item \textbf{Training Steps:} 12 500
    \item \textbf{Dataloader Workers:} 8
    \item \textbf{Batch Size:} 1 (with Gradient Accumulation steps = 4, effective batch size = 4)
    \item \textbf{Loss Weights:}
        \begin{itemize}
            \item Adversarial Loss (\(\lambda_{GAN}\)): 0.5
            \item Cycle Consistency Loss (\(\lambda_{cycle}\)): 1.0
            \item Identity Loss (\(\lambda_{idt}\)): 0.5
        \end{itemize}
\end{itemize}

\subsection{Inference}
Inference was performed using the \verb|a2b| (Source to Target) direction. To enable controllable translation intensity, specifically for modulating fog density and atmospheric effects, we implemented a post-inference linear interpolation strategy. Unlike multi-step diffusion models where "strength" often modulates the number of denoising steps, CycleGAN-Turbo performs image-to-image translation in a single deterministic forward pass. Consequently, intensity control was achieved by computing a pixel-wise weighted average between the original source image (\(I_{source}\)) and the translated output (\(I_{trans}\)). The final output image (\(I_{final}\)) is defined as:

\begin{center}
    \(I_{final}=(1-\alpha)\times I_{source} + \alpha \times I_{trans}\)
\end{center}
where \(\alpha \in [0,1]\) represents the strength parameter. An \(\alpha\) of \(1.0\) yields the full model output, while lower values preserve more of the original scene's signal, effectively simulating variable transparency for atmospheric effects like fog.
\section{Dataset Statistics}
The final curated dataset comprised a total of \textbf{7000} images distributed across four distinct atmospheric domains:
\begin{itemize}
  \item \textbf{day} (2,894 images)
  \item \textbf{sunset/sunrise} (2,283 images)
  \item \textbf{stormy} (dropped)
  \item \textbf{foggy} (1,120 images)
  \item \textbf{night} (703 images)
\end{itemize}
\begin{figure}[h]
    \centering
    \includegraphics[width=0.40\textwidth]{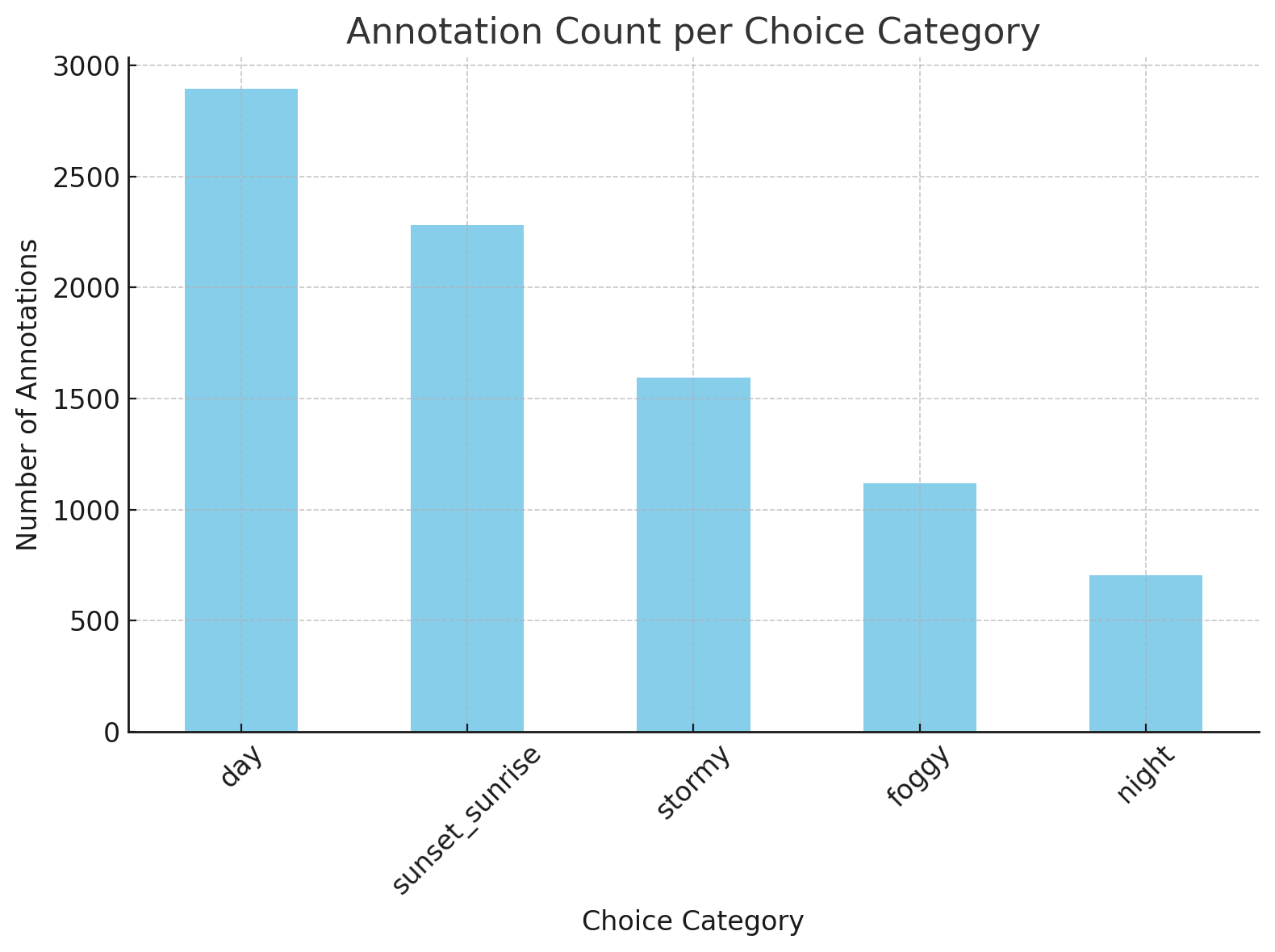}
    \caption{Annotation Count per Choice Category}
    \label{fig:mesh1}
\end{figure}

The class distribution reveals a significant imbalance, with the \verb|day| category (2894 images) outnumbering the \verb|night| category (703 images) by a ratio of approximately \textbf{4.1:1}. This imbalance presents a specific challenge for adversarial training, as the discriminator has fewer examples to learn the target \verb|night| distribution compared to the source \verb|day| distribution.

All images were normalized to a resolution of \(286 \times 286\) for the final dataset. Following the 90/10 partitioning strategy, the images were distributed into training and testing sets as detailed in Table 1.

\begin{table}[h]
    \centering
    \begin{tabular}{cccc}
        \hline
        \textbf{Category} & Total Images & Training Set (90\verb|%|) & Test Set (10\verb|%|)\\
        \hline
        Day & 2894 & 2605 & 289\\
        Sunset/Sunrise & 2283 & 2055 & 228\\
        Foggy & 1120 & 1008 & 112\\
        Night & 703 & 633 & 70\\
        \textbf{Total} & \textbf{7000} & \textbf{6301} & \textbf{699}\\
        \hline
    \end{tabular}
    \caption{Distribution of images across atmospheric domains and data partitions.}
    \label{tab:placeholder}
\end{table}

A qualitative analysis of the \verb|night| category revealed a critical lack of variety; the majority of available imagery depicted coastal or harbor environments heavily illuminated by artificial light sources, such as street lights and pier illumination. This semantic bias towards "coastal night" rather than "open-sea night" in the available data introduced significant artifacts in the generated outputs. Specifically, the model frequently hallucinated light streaks and artificial reflections into deep-sea scenes, as it learned intrinsically associate the "Night" domain with the presence of electric lighting rather than natural darkness.
\section{Results}
This section presents the evaluation of the trained CycleGAN-turbo models across three distinct maritime domain translations: Day-To-Foggy, Day-To-Sunset, and Day-To-Night. We begin by examining the quantitative training dynamics, specifically the convergence of the Cycle Consistency Loss, to validate model learning and stability. Next, we provide a qualitative visual assessment of the generated atmospheric conditions, including a comparative study on the impact of varying inference guidance strengths. Finally, we conclude with a targeted analysis of small object preservation, assessing the models' capability to retain fine details during the image translation process.

\subsection{Training Convergence}
Training stability was monitored via the Cycle Consistency Loss (\(\lambda_{cycle}\)), which measures the L1 difference between the original image and its reconstructed version after a round-trip translation. \cite{esriCycleGAN}

The training process for our image-to-image translation models showed consistent progress across all lighting and weather conditions. All three models initiated training with a starting loss between 7.5 to 8.0, reflecting the initial gap between the source and target domains. As the model optimized, we observed the following patterns:

\begin{table}[h]
    \centering
    \begin{tabular}{ccccc}
        \hline
        \textbf{Model} & SL cycle A & SL cycle B & FL cycle A & FL cycle B \\
        \hline
        day2foggy & \(\sim 8.0\) & \(\sim 7.0\) & \(\sim 1.6\) & \(\sim 1.3\) \\
        day2sunset & \(\sim 7.5\) & \(\sim 7.5\) & \(\sim 1.5\) & \(\sim 1.4\) \\
        day2night & \(\sim 7.8\) & \(\sim\ 6.8\) & \(\sim 1.6\) & \(\sim 1.6\) \\
        \hline
    \end{tabular}
    \caption{Convergence patterns of all three model domains. SL = Starting Loss, FL = Final Loss.}
    \label{tab:placeholder}
\end{table}

In the primary translation cycle (cycle A), all three models converged to a nearly identical loss range of 1.5 to 1.6. This suggest that the initial transformation from a clear day image to a stylized target (Night, Fog, or Sunset) presents a similar level of baseline complexity for the generator.

Significant differences emerged in cycle B (the reconstruction cycle). The day2foggy model achieved the lowest final loss at \(\sim 1.3\), followed closely by day2sunset at \(\sim 1.4\). This indicates that the model found it easier to recover original "day" features from foggy and sunset domains than from the night domain.

The day2night model maintained the highest loss (\(\sim 1.6\)) across both cycles. This difficulty stems from the extreme contrast shifts and heavy shadowing in nighttime images, which often obscure the structural data needed for reconstruction.

The lower final values for day2foggy and day2sunset suggest these models succesfully identified more stable feature mappings, likely aided by a more balanced dataset distribution that prevented the "information loss" seen in the night transitions.

\subsection{Qualitative Evaluation}
To assess the visual and structural preservation of the trained models, we conducted a qualitative evaluation using roundtrip translations. This process involves translating an image from its source domain to the target domain and then reconstructing the source (e.g., Day \(\xrightarrow{}\) Foggy \(\xrightarrow{}\) Day), as well as performing the inverse translation (Foggy \(\xrightarrow{}\) Day \(\xrightarrow{}\) Foggy).

\subsubsection{Model A: Day-to-Foggy Translation}
The day2foggy model demonstrated the highest qualitative performance among the three evaluated networks. In both the day-foggy-day and foggy-day-foggy roundtrip translations (Fig. 2), the model effectively synthesized the atmospheric scattering and contrast reduction typical of marine fog. Crucially, the reconstructed images resemble the original inputs very closely. The high fidelity of the reconstructed maritime structures after being subjected to simulated low-visibility conditions indicates that the model successfully learned the atmospheric style without permanently discarding the underlying semantic content.

\subsubsection{Model B: Day-to-Sunset Translation}
The day2sunset model also fared well, yielding realistic atmospheric translations. As illustrated in the day-sunset-day and sunset-day-sunset roundtrips (Fig. 3), the model accurately captured the warm illumination shifts and high-contrast horizons characteristic of the golden hour. The reconstructed images maintains a strong structural resemblance to the originals, benefiting from the balanced distribution between the Day and Sunset training datasets, which allowed for stable cycle consistency optimization.

\subsubsection{Model C: Day-to-Night Translation}
The day2night model exhibited more complex behavior, highlighting specific vulnerabilities related to dataset bias. While the model successfully compressed the dynamic range to simulate nighttime conditions, the day-night-day roundtrip translation frequently introduced severe visual anomalies. Specifically, the model created "light artifacts", hallucinating city lights and reflections into open-water or cloud regions where no light sources existed in the original daytime image (Fig.4 ).

Conversely, the night-day-night translation revealed hallucination errors during the daytime generation phase. For example, as shown in Fig. 4, the model generated a dense forest along a coastline that, in the original nighttime image, contained dimly lit buildings. These artifacts are a direct consequence of the limited variety and strong coastal bias in the overall training data, forcing the model to intrinsically associate the dark coastline with forests, instead of recreating the dimly lit buildings on the coastline.

\begin{figure*}[t]
    \centering
    \setlength{\tabcolsep}{0pt} 
    
    \begin{tabular}{|c|}
        \hline
        \begin{tabularx}{0.98\textwidth}{CCC}
            \addlinespace[1ex]
            \multicolumn{3}{c}{\textbf{Model A: Day $\leftrightarrow$ Foggy}} \\ \addlinespace[1ex]
            \begin{subfigure}[b]{0.25\textwidth}
                \centering
                \includegraphics[width=0.98\textwidth]{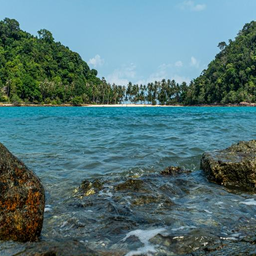}
                \caption{Real Day}
            \end{subfigure} &
            \begin{subfigure}[b]{0.25\textwidth}
                \centering
                \includegraphics[width=0.98\textwidth]{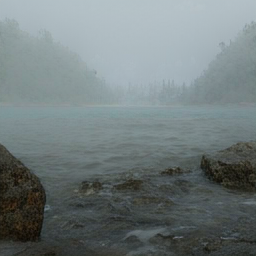}
                \caption{Generated Foggy}
            \end{subfigure} &
            \begin{subfigure}[b]{0.25\textwidth}
                \centering
                \includegraphics[width=0.98\textwidth]{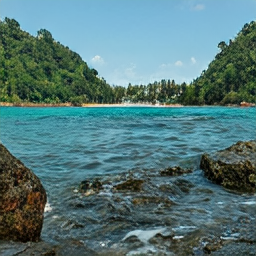}
                \caption{Reconstructed Day}
            \end{subfigure} \\ \addlinespace[2ex]
            
            \begin{subfigure}[b]{0.25\textwidth}
                \centering
                \includegraphics[width=0.98\textwidth]{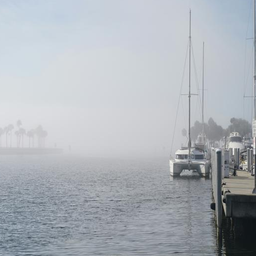}
                \caption{Real Foggy}
            \end{subfigure} &
            \begin{subfigure}[b]{0.25\textwidth}
                \centering
                \includegraphics[width=0.98\textwidth]{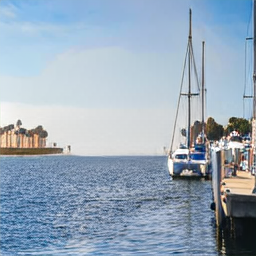}
                \caption{Generated Day}
            \end{subfigure} &
            \begin{subfigure}[b]{0.25\textwidth}
                \centering
                \includegraphics[width=0.98\textwidth]{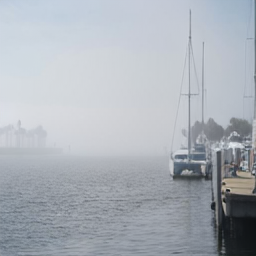}
                \caption{Reconstructed Foggy}
            \end{subfigure} \\ \addlinespace[1ex]
        \end{tabularx} \\ \hline
    \end{tabular}

    \vspace{15pt} 

    \caption{Qualitative results of the full cycle-consistency round trip of the day2foggy model.}
    \label{fig:round_trip_day2foggy}
\end{figure*}

\begin{figure*}[t]
    \centering
    \setlength{\tabcolsep}{0pt} 

    \begin{tabular}{|c|}
        \hline
        \begin{tabularx}{0.98\textwidth}{CCC}
            \addlinespace[1ex]
            \multicolumn{3}{c}{\textbf{Model B: Day $\leftrightarrow$ Sunset}} \\ \addlinespace[1ex]
            \begin{subfigure}[b]{0.25\textwidth}
                \centering
                \includegraphics[width=0.98\textwidth]{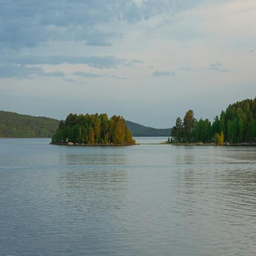}
                \caption{Real Day}
            \end{subfigure} &
            \begin{subfigure}[b]{0.25\textwidth}
                \centering
                \includegraphics[width=0.98\textwidth]{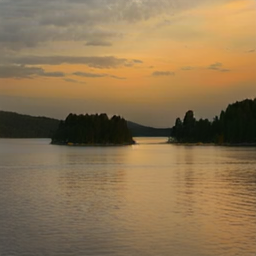}
                \caption{Generated Sunset}
            \end{subfigure} &
            \begin{subfigure}[b]{0.25\textwidth}
                \centering
                \includegraphics[width=0.98\textwidth]{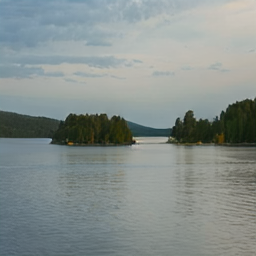}
                \caption{Reconstructed Day}
            \end{subfigure} \\ \addlinespace[2ex]
            
            \begin{subfigure}[b]{0.25\textwidth}
                \centering
                \includegraphics[width=0.98\textwidth]{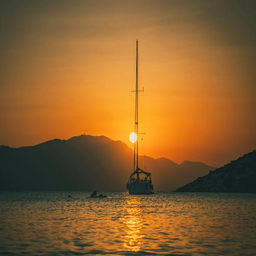}
                \caption{Real Sunset}
            \end{subfigure} &
            \begin{subfigure}[b]{0.25\textwidth}
                \centering
                \includegraphics[width=0.98\textwidth]{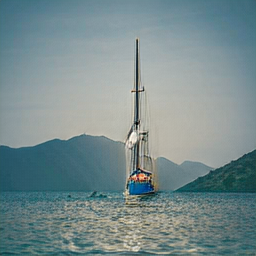}
                \caption{Generated Day}
            \end{subfigure} &
            \begin{subfigure}[b]{0.25\textwidth}
                \centering
                \includegraphics[width=0.98\textwidth]{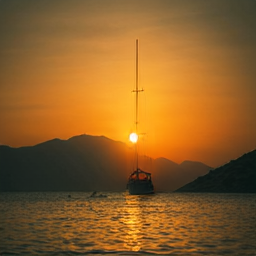}
                \caption{Reconstructed Sunset}
            \end{subfigure} \\ \addlinespace[1ex]
        \end{tabularx} \\ \hline
    \end{tabular}

    \caption{Qualitative results of the full cycle-consistency round trip of the day2sunset model.}
    \label{fig:round_trip_day2sunset}
\end{figure*}

\begin{figure*}[t]
    \centering
    \setlength{\tabcolsep}{0pt} 

    \begin{tabular}{|c|}
        \hline
        \begin{tabularx}{0.98\textwidth}{CCC}
            \addlinespace[1ex]
            \multicolumn{3}{c}{\textbf{Model C: Day $\leftrightarrow$ Night}} \\ \addlinespace[1ex]
            \begin{subfigure}[b]{0.25\textwidth}
                \centering
                \includegraphics[width=0.98\textwidth]{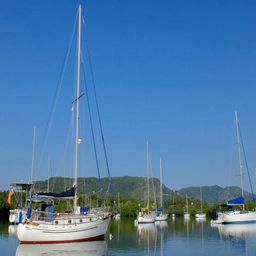}
                \caption{Real Day}
            \end{subfigure} &
            \begin{subfigure}[b]{0.25\textwidth}
                \centering
                \includegraphics[width=0.98\textwidth]{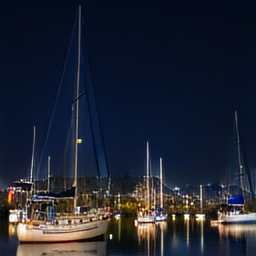}
                \caption{Generated Night}
            \end{subfigure} &
            \begin{subfigure}[b]{0.25\textwidth}
                \centering
                \includegraphics[width=0.98\textwidth]{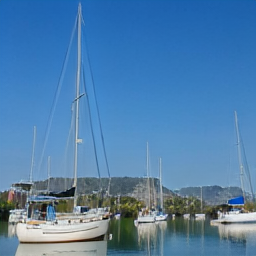}
                \caption{Reconstructed Day}
            \end{subfigure} \\ \addlinespace[2ex]
            
            \begin{subfigure}[b]{0.25\textwidth}
                \centering
                \includegraphics[width=0.98\textwidth]{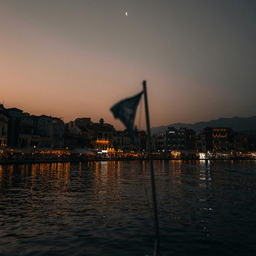}
                \caption{Real Night}
            \end{subfigure} &
            \begin{subfigure}[b]{0.25\textwidth}
                \centering
                \includegraphics[width=0.98\textwidth]{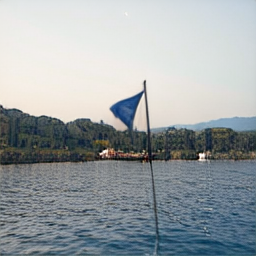}
                \caption{Generated Day}
            \end{subfigure} &
            \begin{subfigure}[b]{0.25\textwidth}
                \centering
                \includegraphics[width=0.98\textwidth]{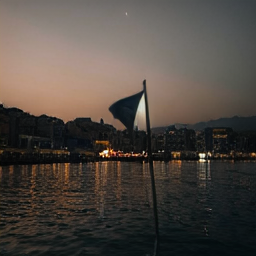}
                \caption{Reconstructed Night}
            \end{subfigure} \\ \addlinespace[1ex]
        \end{tabularx} \\ \hline
    \end{tabular}

    \caption{Qualitative results of the full cycle-consistency round trip of the day2night model.}
    \label{fig:round_trip_day2night}
\end{figure*}

\subsection{Inference Parameters Study}
To determine the optimal balance between the intensity of the applied atmospheric condition and the structural preservation of the original image, we conducted an inference parameter study. The guidance strength parameter was evaluated across all three models at values of 0.7, 0.9, and 1.0 (full strength), as illustrated in Fig. 5. Preliminary testing indicated that utilizing strength values lower than 0.7 caused the original daytime features to remain too prominent, resulting in unnatural composite that failed to convincingly represent target weather/time-of-day domain.

\subsubsection{Model A: Day-to-Foggy}
The day2foggy model proved to be the most highly responsive to guidance strength tuning. As seen in Fig. 5 (b-d), the parameter offers granular control over the simulated visibility. At a strength of 0.7, the model applies a light haze, partially obscuring the background tree line while maintaining the crispness of the foreground motorboat. Increasing the strength to 0.9 and subsequently to 1.0 increases the simulated atmospheric scattering. At full strength (1.0), the background coastline is entirely engulfed in fog, while the structural details of the vessel itself remain well-preserved.

\subsubsection{Model B: Day-to-Sunset}
Conversely, the day2sunset model exhibited significantly less variance across the tested strength values. At a strength of 0.7 (Fig. 5f), the model already effectively shifts the global illumination to warm, golden-hour tones and introduces sunset reflections on the water's surface. Pushing the strength to 0.9 and 1.0 (Fig. 5g, 5h) slightly deepens the orange hues of the sky and introduces a localized, bright sun near the horizon. However, the overall aesthetic difference between 0.7 and full strength is marginal.

\subsubsection{Model C: Day-to-Night}
The day2night model similarly showed diminishing returns at higher guidance strengths, while also highlighting the model's dataset biases. At 0.7 (Fig. 5j), the image is succesfully darkened, converting the sky deep blue and compressing the dynamic range. However, as the strength is increased to 0.9 and 1.0 (Fig. 5k, 5l) the primary change is not a deeper global darkness, but rather an intensification of the "light artifacts" discussed in section 5.A. At full strength, the model aggressively hallucinates bright city lights on the coastline, complete with light streaks reflecting across the water. \newline

In summary, while the strength parameter provides excellent utility for fine-tuning fog density in the day2foggy model, the day2sunset and day2night models achieve their primary stylistic transformations at around 0.7. Pushing these latter two models to full strength yields minimal structural benefits and, in the case of the nighttime model, actively amplifies unwanted dataset-induced artifacts.

\begin{figure*}[t]
    \centering
    \setlength{\tabcolsep}{0pt} 

    \begin{tabular}{|c|}
        \hline
        \begin{tabularx}{0.98\textwidth}{CCCC}
            \addlinespace[1ex]
            \multicolumn{4}{c}{\textbf{Model A: day2foggy 256x256}} \\ \addlinespace[1ex]
            \begin{subfigure}[b]{0.235\textwidth}
                \centering
                \includegraphics[width=0.98\textwidth]{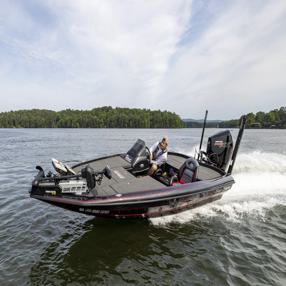}
                \caption{Original}
            \end{subfigure} &
            \begin{subfigure}[b]{0.235\textwidth}
                \centering
                \includegraphics[width=0.98\textwidth]{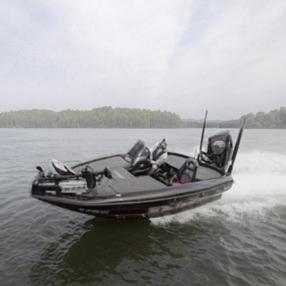}
                \caption{0.7}
            \end{subfigure} &
            \begin{subfigure}[b]{0.235\textwidth}
                \centering
                \includegraphics[width=0.98\textwidth]{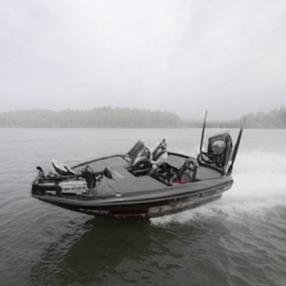}
                \caption{0.9}
            \end{subfigure} &
            \begin{subfigure}[b]{0.235\textwidth}
                \centering
                \includegraphics[width=0.98\textwidth]{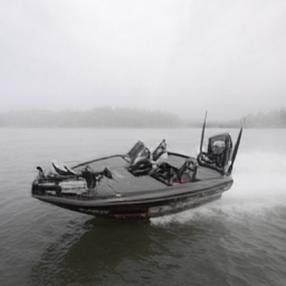}
                \caption{1.0}
            \end{subfigure} \\ \addlinespace[1ex]
        \end{tabularx} \\ \hline
    \end{tabular}

    \vspace{10pt} 

    \begin{tabular}{|c|}
        \hline
        \begin{tabularx}{0.98\textwidth}{CCCC}
            \addlinespace[1ex]
            \multicolumn{4}{c}{\textbf{Model B: day2sunset 256x256}} \\ \addlinespace[1ex]
            \begin{subfigure}[b]{0.235\textwidth}
                \centering
                \includegraphics[width=0.98\textwidth]{images/qualitative/original.jpg}
                \caption{Original}
            \end{subfigure} &
            \begin{subfigure}[b]{0.235\textwidth}
                \centering
                \includegraphics[width=0.98\textwidth]{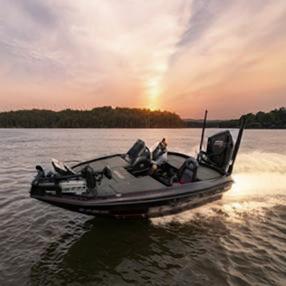}
                \caption{0.7}
            \end{subfigure} &
            \begin{subfigure}[b]{0.235\textwidth}
                \centering
                \includegraphics[width=0.98\textwidth]{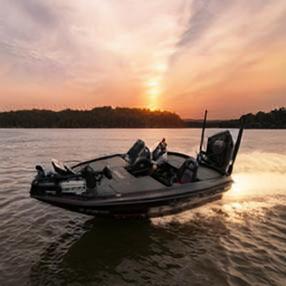}
                \caption{0.9}
            \end{subfigure} &
            \begin{subfigure}[b]{0.235\textwidth}
                \centering
                \includegraphics[width=0.98\textwidth]{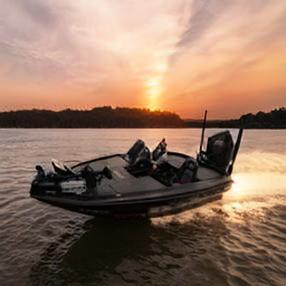}
                \caption{1.0}
            \end{subfigure} \\ \addlinespace[1ex]
        \end{tabularx} \\ \hline
    \end{tabular}

    \vspace{10pt}

    \begin{tabular}{|c|}
        \hline
        \begin{tabularx}{0.98\textwidth}{CCCC}
            \addlinespace[1ex]
            \multicolumn{4}{c}{\textbf{Model C: day2night 256x256}} \\ \addlinespace[1ex]
            \begin{subfigure}[b]{0.235\textwidth}
                \centering
                \includegraphics[width=0.98\textwidth]{images/qualitative/original.jpg}
                \caption{Original}
            \end{subfigure} &
            \begin{subfigure}[b]{0.235\textwidth}
                \centering
                \includegraphics[width=0.98\textwidth]{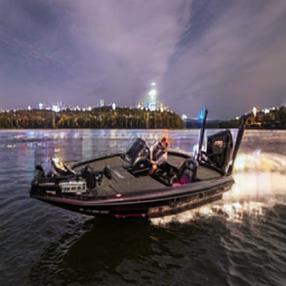}
                \caption{0.7}
            \end{subfigure} &
            \begin{subfigure}[b]{0.235\textwidth}
                \centering
                \includegraphics[width=0.98\textwidth]{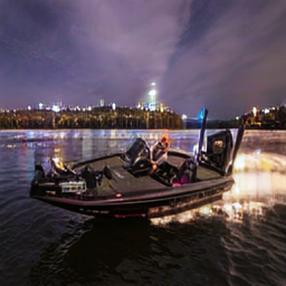}
                \caption{0.9}
            \end{subfigure} &
            \begin{subfigure}[b]{0.235\textwidth}
                \centering
                \includegraphics[width=0.98\textwidth]{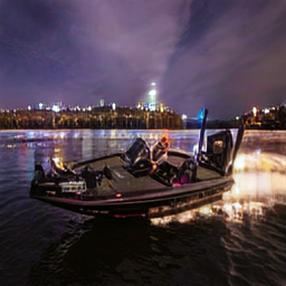}
                \caption{1.0}
            \end{subfigure} \\ \addlinespace[1ex]
        \end{tabularx} \\ \hline
    \end{tabular}

    \caption{Visual comparison across different maritime translation models at variable strengths. 1.0 represents full strength.}
    \label{fig:full_comparison}
\end{figure*}

\subsection{Small Object Preservation Analysis}

\begin{figure*}[t]
    \centering
    \setlength{\tabcolsep}{0pt} 

    \begin{tabular}{|c|}
        \hline
        \begin{tabularx}{0.98\textwidth}{CCC}
            \addlinespace[1ex]
            \multicolumn{3}{c}{\textbf{Small Object Retention Comparison: Day $\rightarrow$ Sunset}} \\ \addlinespace[1ex]
            \hline
            \textbf{Original Day} & \textbf{CycleGAN-turbo} & \textbf{HiDT} \\ \addlinespace[0.5ex]
            \hline
            
            \begin{subfigure}[b]{0.31\textwidth}
                \centering
                \includegraphics[width=0.98\textwidth]{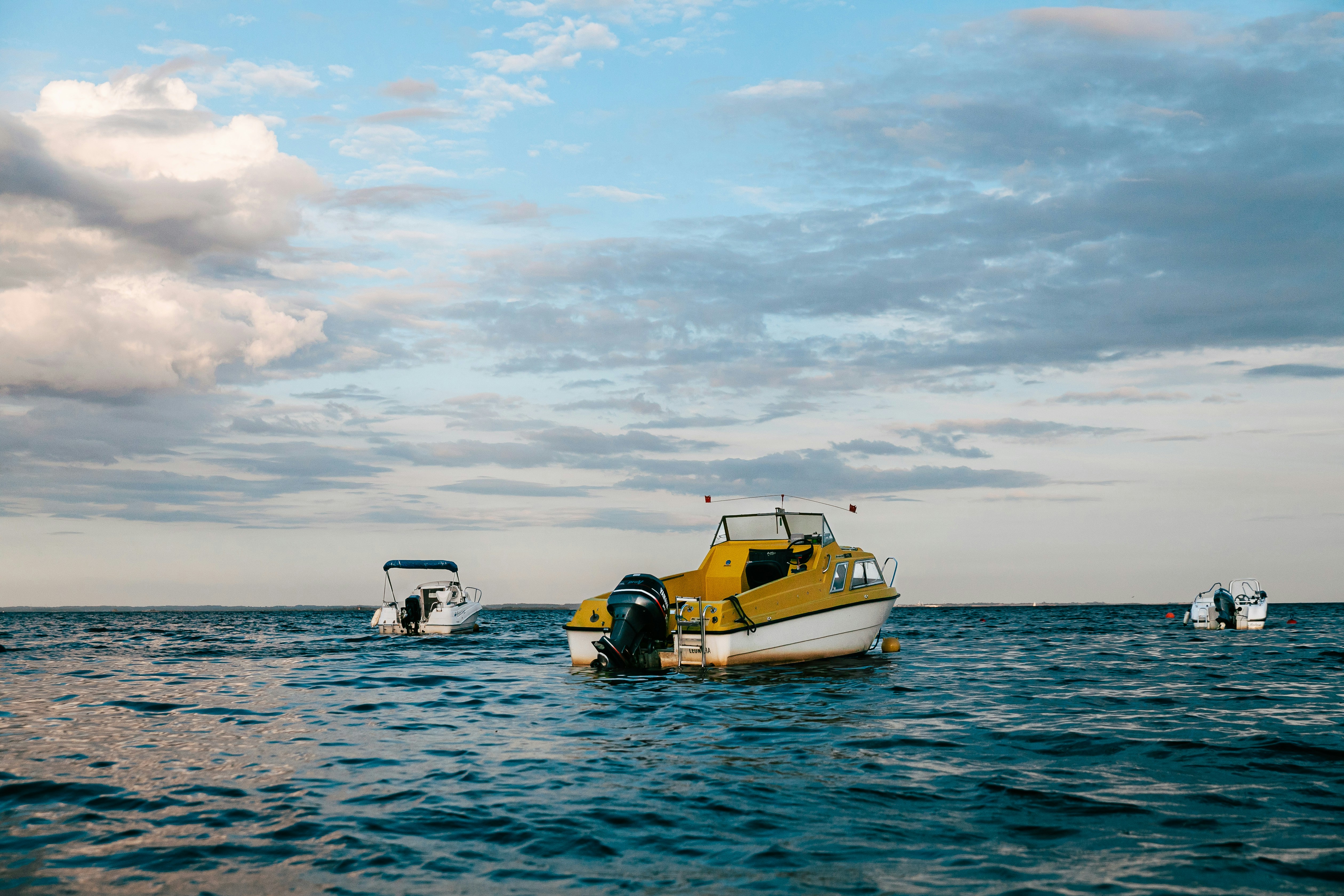}
                \caption{Original Image}
            \end{subfigure} &
            \begin{subfigure}[b]{0.31\textwidth}
                \centering
                \includegraphics[width=0.98\textwidth]{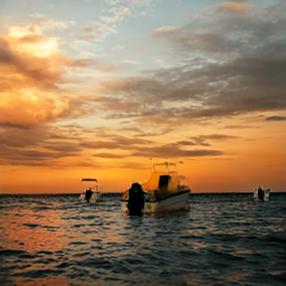}
                \caption{day2sunset}
            \end{subfigure} &
            \begin{subfigure}[b]{0.31\textwidth}
                \centering
                \includegraphics[width=0.98\textwidth]{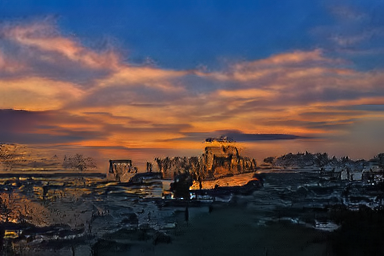}
                \caption{5minute}
            \end{subfigure} \\ \addlinespace[2ex]

            \begin{subfigure}[b]{0.31\textwidth}
                \centering
                \includegraphics[width=0.98\textwidth]{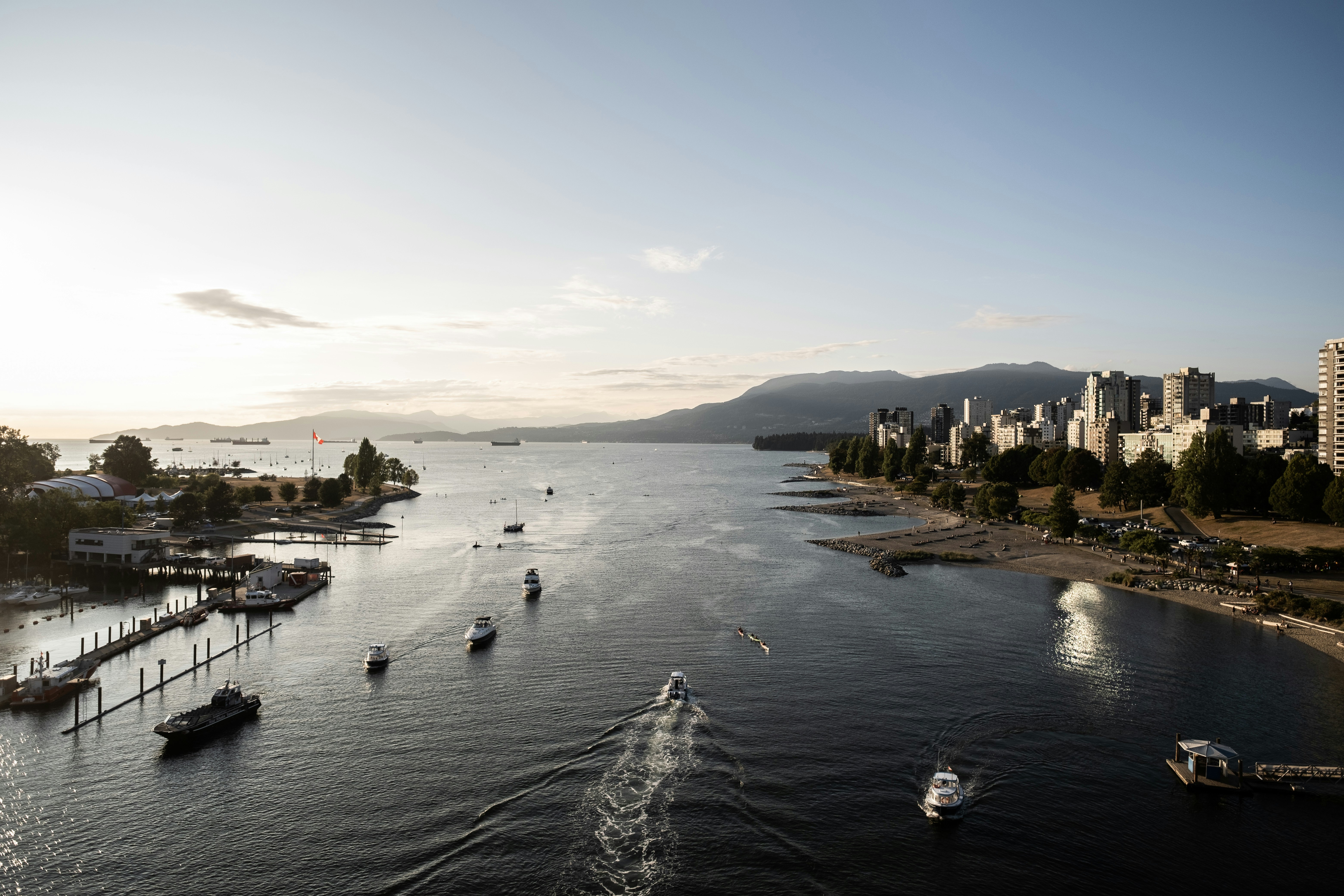}
                \caption{Original Image}
            \end{subfigure} &
            \begin{subfigure}[b]{0.31\textwidth}
                \centering
                \includegraphics[width=0.98\textwidth]{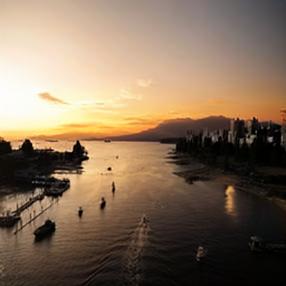}
                \caption{day2sunset}
            \end{subfigure} &
            \begin{subfigure}[b]{0.31\textwidth}
                \centering
                \includegraphics[width=0.98\textwidth]{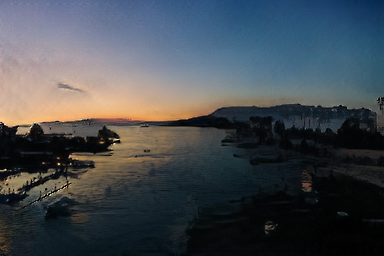}
                \caption{5minute}
            \end{subfigure} \\ \addlinespace[1ex]

            \begin{subfigure}[b]{0.31\textwidth}
                \centering
                \includegraphics[width=0.98\textwidth]{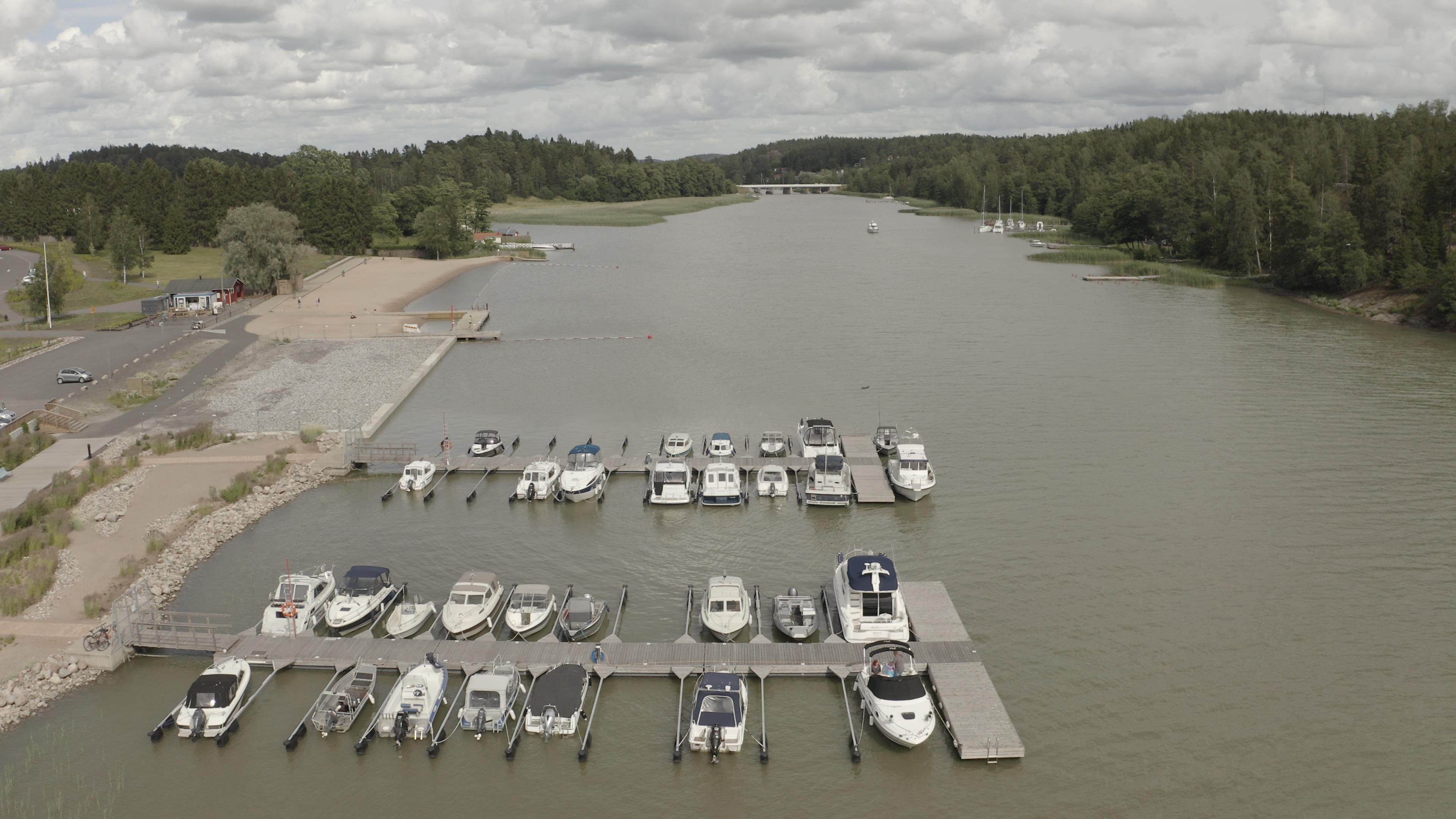}
                \caption{Original Image}
            \end{subfigure} &
            \begin{subfigure}[b]{0.31\textwidth}
                \centering
                \includegraphics[width=0.98\textwidth]{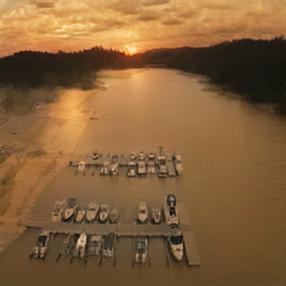}
                \caption{day2sunset}
            \end{subfigure} &
            \begin{subfigure}[b]{0.31\textwidth}
                \centering
                \includegraphics[width=0.98\textwidth]{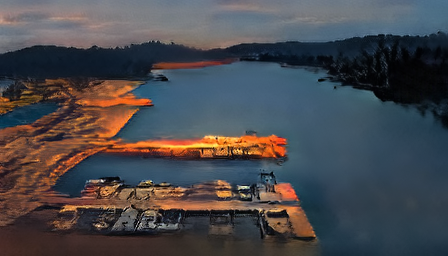}
                \caption{5minute}
            \end{subfigure} \\ \addlinespace[1ex]
        \end{tabularx} \\ \hline
    \end{tabular}

    \caption{Comparative analysis between CycleGAN-turbo and HiDT for Day-to-Sunset translation.}
    \label{fig:model_comparison_day_sunset}
\end{figure*}

\begin{figure*}[t]
    \centering
    \setlength{\tabcolsep}{0pt} 
    
    \begin{tabular}{|c|}
        \hline
        \begin{tabularx}{0.98\textwidth}{CCC}
            \addlinespace[1ex]
            \multicolumn{3}{c}{\textbf{Small Object Retention Comparison: Day $\rightarrow$ Night}} \\ \addlinespace[1ex]
            \hline
            \textbf{Original Day} & \textbf{CycleGAN-turbo} & \textbf{HiDT} \\ \addlinespace[0.5ex]
            \hline
            
            \begin{subfigure}[b]{0.31\textwidth}
                \centering
                \includegraphics[width=0.98\textwidth]{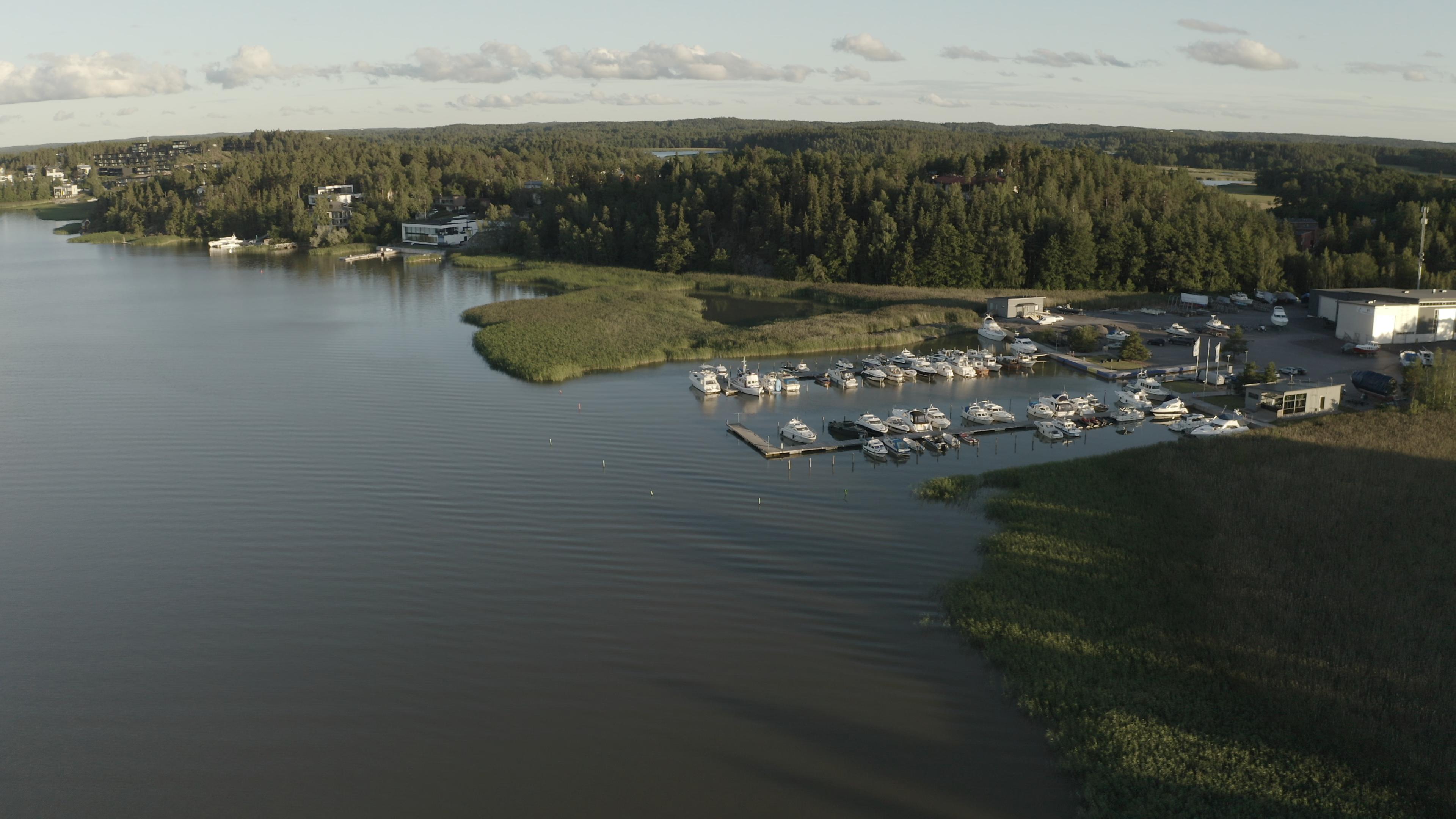}
                \caption{Original Image}
            \end{subfigure} &
            \begin{subfigure}[b]{0.31\textwidth}
                \centering
                \includegraphics[width=0.98\textwidth]{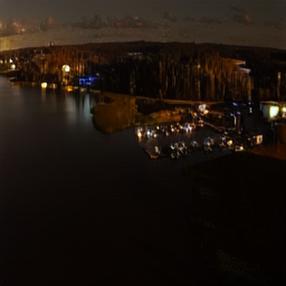}
                \caption{day2night}
            \end{subfigure} &
            \begin{subfigure}[b]{0.31\textwidth}
                \centering
                \includegraphics[width=0.98\textwidth]{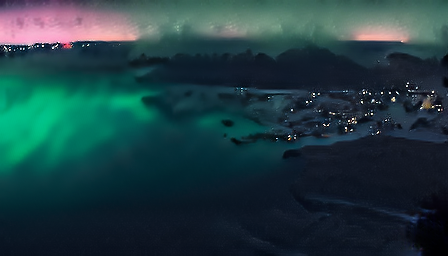}
                \caption{another night}
            \end{subfigure} \\ \addlinespace[2ex]

            \begin{subfigure}[b]{0.31\textwidth}
                \centering
                \includegraphics[width=0.98\textwidth]{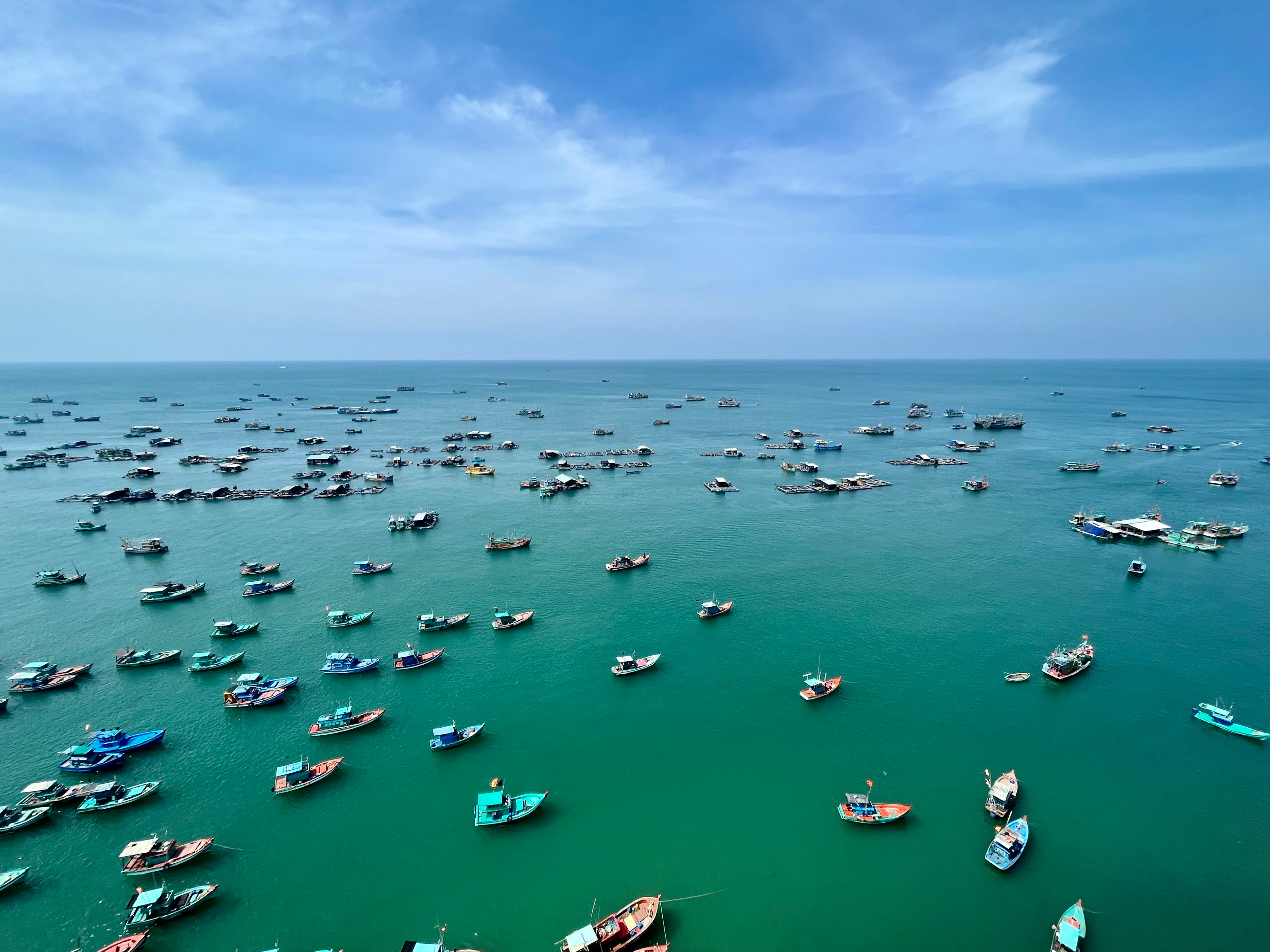}
                \caption{Original Image}
            \end{subfigure} &
            \begin{subfigure}[b]{0.31\textwidth}
                \centering
                \includegraphics[width=0.98\textwidth]{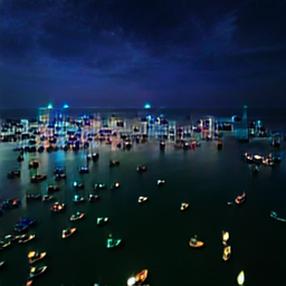}
                \caption{day2night}
            \end{subfigure} &
            \begin{subfigure}[b]{0.31\textwidth}
                \centering
                \includegraphics[width=0.98\textwidth]{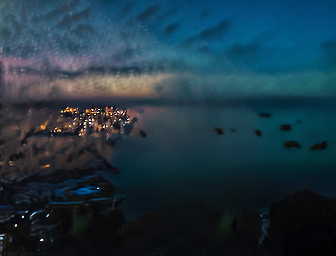}
                \caption{another night}
            \end{subfigure} \\ \addlinespace[1ex]

            \begin{subfigure}[b]{0.31\textwidth}
                \centering
                \includegraphics[width=0.98\textwidth]{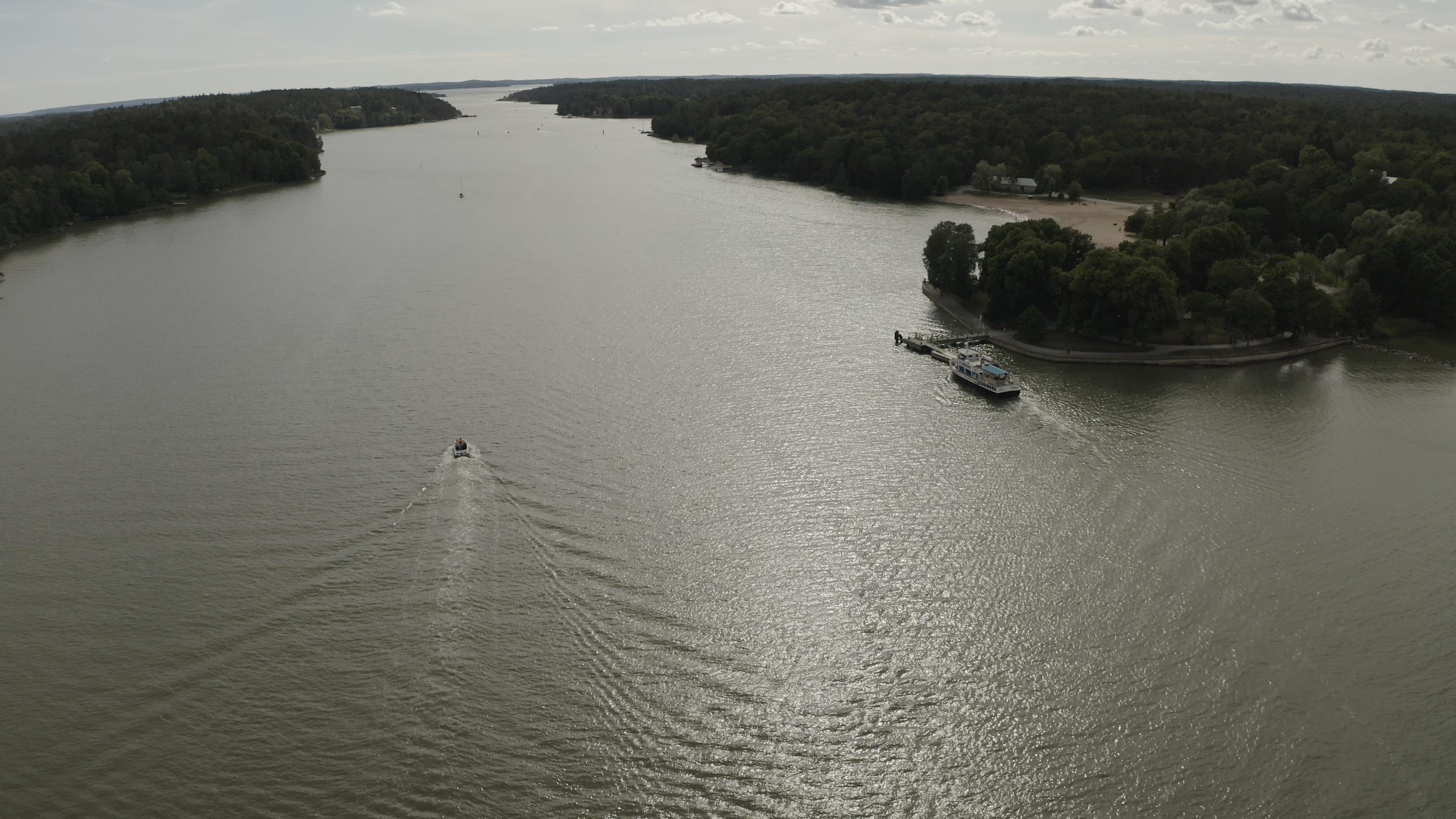}
                \caption{Original Image}
            \end{subfigure} &
            \begin{subfigure}[b]{0.31\textwidth}
                \centering
                \includegraphics[width=0.98\textwidth]{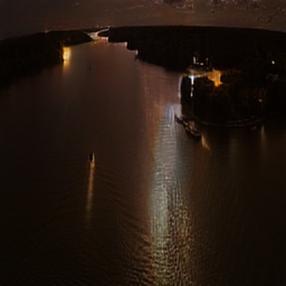}
                \caption{day2night}
            \end{subfigure} &
            \begin{subfigure}[b]{0.31\textwidth}
                \centering
                \includegraphics[width=0.98\textwidth]{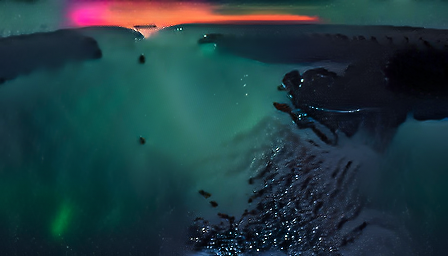}
                \caption{another night}
            \end{subfigure} \\ \addlinespace[1ex]
        \end{tabularx} \\ \hline
    \end{tabular}

    \caption{Comparative analysis between CycleGAN-turbo and HiDT for Day-to-Night translation.}
    \label{fig:model_comparison_grid}
\end{figure*}

To evaluate our model's capacity to retain critical navigational features, such as distant vessels, sea marks, and fine structural geometries, we conducted a comparative analysis focusing specifically on small object preservation. We benchmark our CycleGAN-turbo models against the High-Resolution Daytime Translation (HiDT) framework \cite{anokhin2020hidt}, a well-established baseline previously utilized for synthetic data augmentation.

\subsubsection{Architectural Differences: HiDT vs. CycleGAN-turbo}
The foundational difference between the two models lies in their architectural approaches to balancing global atmospheric style transfer with local detail retention. HiDT operates by decomposing input images into separate content and style representations. These representations are then combined using Adaptive Instance Normalization (AdaIN) layers within an encoder-decoder architecture. \cite{anokhin2020hidt} To preserve fine details without "leaking" the original image's style into the generated output, HiDT employs skip connections that are heavily modulated by an additional AdaIN block. \cite{anokhin2020hidt} While effective for general landscape translation, this AdaIN-modulated feature combination can sometimes lead to the washing out or blurring of minute, high-contrast edges of small maritime objects.

In contrast, our approach utilizes the CycleGAN-turbo architecture, which integrates cycle-consistency constraints with semantic priors of a pre-trained, one-step Latent Diffusion Model. A known vulnerability of standard latent diffusion is the Variational Autoencoder (VAE) bottleneck; the image encoder spatially compresses inputs (e.g., by a factor of 8), frequently causing small objects to be discarded as high-frequency noise during the compression phase. To overcome this and explicitly preserve small object fidelity, CycleGAN-turbo incorporates zero-convolution skip connections. \cite{parmar2024cycleganturbo} These connections establish a direct pathway from the downsampling encoder blocks to the upsampling decoder blocks, effectively bypassing the latent compression bottleneck. As a result, the high-frequency pixel details defining small maritime objects are forcibly retained and integrated with the newly synthesized atmospheric noise map.

\subsubsection{Analysis}
The day2sunset model demonstrated the highest overall performance in our experiments, which is directly reflected in its superior small object preservation. In the evaluated scenes, the CycleGAN-turbo architecture consistently maintained the structural integrity of minute details. Conversely, while the HiDT baseline achieved decent global color shift in certain scenarios (Fig. 6f), it consistently blurred small objects, rendering them unrecognizable. In the first scene, HiDT failed to accurately generate water textures, resulting in a blurry rendering of the waves, whereas our model preserved the dynamic texture of the sea surface. This disparity is particularly evident in the third scene (Fig. 6h), which depicts a pier with parked boats, The CycleGAN-turbo model retained the complex structural outlines of the pier and the individual boat, whereas the HiDT translation degraded into a blurry output where most vessels lost their defining geometric characteristics.

Although the day2night model was our least performant model overall due to the previously discussed light artifacts due to dataset imbalance, it still outperformed HiDT in terms of object and structural preservation. The extreme dynamic range compression required for nighttime translation proved highly challenging for the baseline model. In the first scene (Fig. 7a-c), a distant drone shot of a pier with a visible horizon, the boats are too far away for either model to render perfectly. However, the general structural composition of the scene is much better preserved by the CycleGAN-turbo model. The second scene (Fig. 7d-f), featuring dozens of boats on the open sea, highlights a vast performance gap; our day2night model successfully retained the fleet, while HiDT generated an image so heavily blurred that the semantic context of the scene was completely lost. Similarly, in the third scene (Fig. 7g-i), the CycleGAN-turbo output is appropriately dark, and while a small boat in the frame is reduced to a dark spot, it accurately retains its original size and overall shape. In contrast, the HiDT output is again overwhelmingly blurred and fails to portray the original scene.
\\
Beyond small object preservation, out visual analysis revealed that the HiDT framework struggles fundamentally with maritime environments, even when clear horizons and landmasses are visible. Furthermore, the HiDT's night generation exhibits exhibits a dataset bias; the outputs are frequently infused with vibrant, unnatural colors, heavily resembling the aurora borealis in both the sky and water reflections. Finally, HiDT's detail synthesis is inconsistent. It occasionally generates a small, localized patch of high-detail textures (such as a decently accurate wave pattern) while reducing the remainder of the image to an unrecognizable blur. These observations underscore the limitations of using heavily modulated feature recombination for dynamic, unstructured maritime domains, further validating the necessity of our zero-convolution diffusion approach.

\section{Conclusion}
In this paper, we presented a framework for generating synthetic maritime training data to address the critical data scarcity bottleneck in autonomous surface vessel navigation. By adapting the one-step CycleGAN-turbo architecture, we established an efficient, unpaired image-to-image translation pipeline capable of synthesizing adverse weather and lighting conditions from standard daytime imagery. Crucially, our approach succesfully mitigates the spatial compression bottleneck inherent in standard latent diffusion models. By incorporating zero-convolution skip connections, our models explicitly preserve the fine structural details and geometric boundaries of small, navigationally critical objects, a task where baseline models like HiDT consistently suffer from severe blurring and feature loss.

Our evaluations demonstrated that the day2foggy and day2sunset models achieve high visual fidelity and structural retention. Furthermore, our inference parameter study highlighted the framework's flexibility, particularly in offering granural control over simulated fog density without degrading the underlying semantic content. While the day2night model succesfully compressed dynamic ranges, it also underscored a vital limitation: the susceptibility of generative models semantic hallucinations. The generation of artificial coastal light streaks in open-water scenes directly highlights the necessity for rigorous dataset curation to eliminate semantic biases in future iterations.

Ultimately, this research provides a robust, structure-aware data synthesis tool that significantly reduces the reliance in physically collecting rare maritime data. By proving that synthetic augmentation can retain the minute details required for accurate object detection, this framework paves the way for the development of resilient, all-weather perception systems for maritime autonomy.
\section{Acknowledgements}

We extend our gratitude to the Uncrewed Surface Vessels for Automated Critical Infrastructure Protection (USVA) project funded by Business Finland for their support of this research.

\bibliographystyle{ieeetr}
\bibliography{mybibfile}

\end{document}